\documentclass[letterpaper, 10 pt, journal, twoside]{./IEEEtran}
\IEEEoverridecommandlockouts

\usepackage[utf8]{inputenc}
\usepackage[english]{babel}
\usepackage[T1]{fontenc}
\usepackage{amssymb,amsfonts}
\usepackage[cmex10]{amsmath}
\usepackage{dsfont}
\usepackage{algorithm}
\usepackage{algpseudocode}
\usepackage{multirow}
\usepackage{graphicx}
\usepackage{textcomp}
\usepackage{xcolor}
\usepackage[binary-units=true]{siunitx}
\usepackage{tikz}
\usepackage[caption=false,font=footnotesize]{subfig}
\usepackage{hyperref}

\newcommand{\disable}[1]{}
\newcommand{\Tmax}{\ensuremath{\text{C}\textsubscript{max}}}
\newcommand{\Dmax}{\ensuremath{\text{D}\textsubscript{max}}}
\newcommand{\Cmax}{\ensuremath{\text{C}\textsubscript{max}}}
\newcommand{\Vmax}{\ensuremath{\text{v}\textsubscript{max}}}
\newcommand{\Vinsp}{\ensuremath{\text{v}\textsubscript{insp}}}
\newcommand{\Amax}{\ensuremath{\text{a}\textsubscript{max}}}
\newcommand{\matr}[1]{\mathbf{#1}}

\newcommand{\boundellipse}[3]
{(#1) ellipse (#2 and #3)
}

\newcommand{\reviewi}[1]{{#1}} 
\newcommand{\reviewii}[1]{{#1}} 
\newcommand{\reviewiii}[1]{{#1}} 
\newcommand{\reviewiv}[1]{{#1}} 
\newcommand{\reviewv}[1]{{#1}} 
\newtheorem{problem}{Problem}[section]
\usepackage{scalerel}
\usetikzlibrary{svg.path}
\definecolor{orcidlogocol}{HTML}{A6CE39}
\tikzset{
orcidlogo/.pic={
  \fill[orcidlogocol] svg{M256,128c0,70.7-57.3,128-128,128C57.3,256,0,198.7,0,128C0,57.3,57.3,0,128,0C198.7,0,256,57.3,256,128z};
  \fill[white] svg{M86.3,186.2H70.9V79.1h15.4v48.4V186.2z}
               svg{M108.9,79.1h41.6c39.6,0,57,28.3,57,53.6c0,27.5-21.5,53.6-56.8,53.6h-41.8V79.1z M124.3,172.4h24.5c34.9,0,42.9-26.5,42.9-39.7c0-21.5-13.7-39.7-43.7-39.7h-23.7V172.4z}
               svg{M88.7,56.8c0,5.5-4.5,10.1-10.1,10.1c-5.6,0-10.1-4.6-10.1-10.1c0-5.6,4.5-10.1,10.1-10.1C84.2,46.7,88.7,51.3,88.7,56.8z};
}
}
\newcommand\orcidicon[1]{\href{https://orcid.org/#1}{\mbox{\scalerel*{
\begin{tikzpicture}[yscale=-1,transform shape]
\pic{orcidlogo};
\end{tikzpicture}
}{|}}}}

\begin{document}
\bstctlcite{mstsp-bib}

\title{Multi-tour Set Traveling Salesman Problem in Planning Power Transmission Line Inspection}

\author{František Nekovář$^{\orcidicon{0000-0002-1975-078X}}$, \and Jan Faigl$^{\orcidicon{0000-0002-6193-0792}}$, \and Martin Saska$^{\orcidicon{0000-0001-7106-3816}}$
\thanks{This work was supported by the Czech Science Foundation (GAČR) under research
project No. 19-20238S, the European Unions Horizon 2020 research and innovation  programme AERIAL-CORE under grant agreement no. 871479, and the CTU grant no. SGS20/174/OHK3/3T/13.
Computational resources were supplied by the project ``e-Infrastruktura CZ'' (e-INFRA LM2018140) provided within the program Projects of Large Research, Development and Innovations Infrastructures.}
\thanks{Authors are with the Czech Technical University, Faculty of Electrical Engineering, Technicka 2, 166 27, Prague, Czech Republic, email: {\tt\small \{nekovfra|faiglj|saskam1\}@fel.cvut.cz}.}
\thanks{Digital Object Identifier (DOI): 10.1109/LRA.2021.3091695.}
}

\maketitle

\begin{abstract}
   This paper concerns optimal power transmission line inspection formulated as a proposed generalization of the traveling salesman problem for a multi-route one-depot scenario.
The problem is formulated for an inspection vehicle with a limited travel budget.
Therefore, the solution can be composed of multiple runs to provide full coverage of the given power lines.
Besides, the solution indicates how many vehicles can perform the inspection in a single run.
The optimal solution of the problem is solved by the proposed Integer Linear Programming (ILP) formulation, which is, however, very computationally demanding.
Therefore, the computational requirements are addressed by the combinatorial metaheuristic.
The employed greedy randomized adaptive search procedure is significantly less demanding while providing competitive solutions and scales better with the problem size than the ILP-based approach.
The proposed formulation and algorithms are demonstrated in a real-world scenario to inspect power line segments at the electrical substation.

\end{abstract}


\section{Introduction}

\begin{figure}[htb]
   \includegraphics[width=\columnwidth]{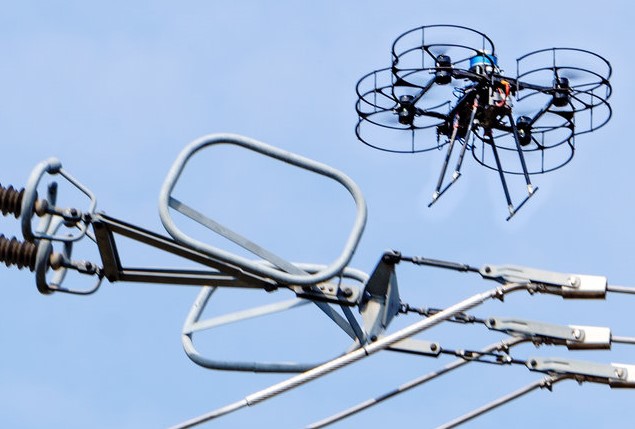}
   \caption{UAV performing visual inspection on a power transmission line.\label{fig:uav_power_line_inspection}}
\end{figure}

\IEEEPARstart{I}{n} this paper, we address the inspection planning of the \emph{Power Transmission Line} (PTL) formulated as a proposed generalization of the well-known combinatorial \emph{Traveling Salesman Problem} (TSP).
The presented method is motivated by the inspection of transmission infrastructure using \emph{Unmanned Aerial Vehicles} (UAVs). 
It is becoming a standard approach advertised as a cost-saving solution compared to either manned inspections or those done using piloted helicopters, such as seen in Fig.~\ref{fig:uav_power_line_inspection}.
Although the development of autonomous inspection is currently in progress, the necessity of having a prepared plan for the efficient utilization of single or multiple vehicles to achieve cost-efficient inspection is desirable for human-piloted UAVs, as well.
The problem addressed in this paper is a major aspect of the objectives of the EU-funded AERIAL CORE project.

The transmission line inspection is performed during vehicle flights close to the power lines.
It is advantageous to perform such an inspection flight between segments of the power lines defined by the pylons~\cite{mendenez16vision}.
Hence, the problem is defined as coverage to visit the transmission line segments, where each segment is a target to be visited.
In the inspection, the segments are organized into a sequence of segments; thus, we formulate the sequencing problem as a variant of the TSP.

A solution of the TSP is a sequence of visits to the given targets that is further generalized to targets grouped into sets, where only a single visit to some target in a set is required, but all sets have to be visited.
Furthermore, within the context of a small UAV with limited operational time, it might not be possible to inspect all the sets in a single run.
Therefore, we generalize the standard TSP formulation to consider limited maximal flight time \Tmax{}, to consider multiple vehicles to cover all segments or a single vehicle with multiple tours, and with the start and end of the respective tours at the common depot.
Such a generalized formulation of the TSP is proposed to describe the herein addressed PTL inspection planning problem, where a UAV has to visit each segment of the PTL exactly once, but in an arbitrary direction.
The travel cost to the particular start of the segment inspection is also relevant as it is part of the travel budget \Tmax{}.
Note, the difference between using a single UAV in a multi-tour scenario and multiple UAVs simultaneously is manifested in the problem cost function.
Since we aim to minimize the total inspection time with the one-vehicle multiple-tours case, minimizing the min-sum of the tour costs is more relevant rather than minimizing the maximum cost from all the tour costs.

\begin{figure}[!htb]
   \includegraphics[width=\columnwidth]{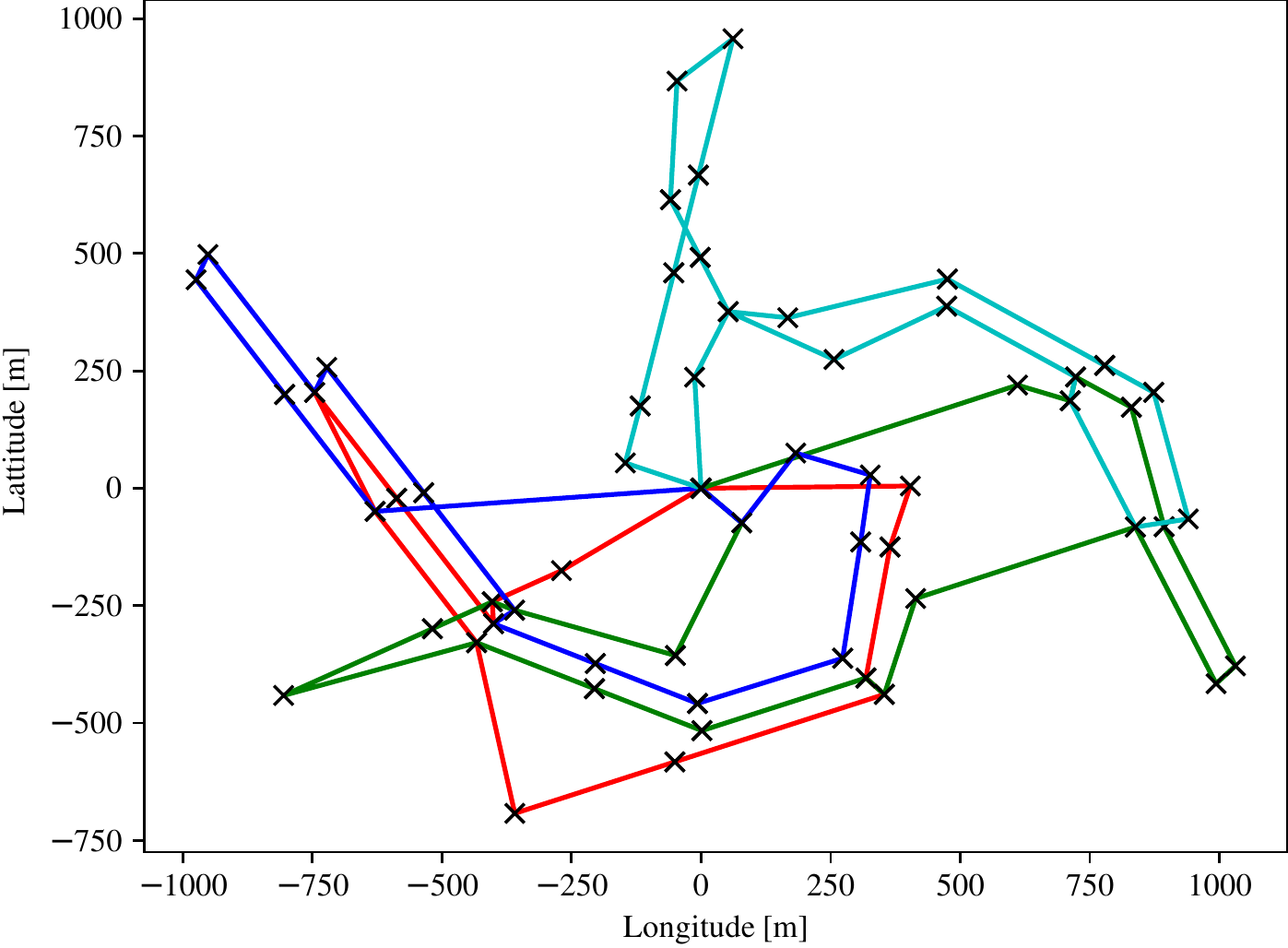}
   \caption{An example of the proposed GRASP-based solution with four UAV routes for inspection of power transmissions lines around Nechranice power line substation, located at 50°20'44.7"N 13°19'30.9"W.}\label{fig:sol_map}
\end{figure}

The proposed TSP-based formulation for PTL inspection is addressed by \emph{Integer Linear Programming} (ILP) to find an optimal solution. However, it is computationally demanding and does not scale with the problem size.
Therefore, we propose a heuristic solution based on the combinatorial metaheuristic \emph{Greedy Randomized Adaptive Search Procedures} (GRASP) to provide a solution in a reasonable computational time, as it is possible that targets could change and a solution would need to be computed on-site.
Based on the performed empirical evaluation, the proposed heuristic solver provides adequate solutions with significantly lower computational requirements than the optimal ILP.
An example of the solution found is depicted in Fig.~\ref{fig:sol_map}.
Our contribution is the formulation of the PTL inspection as the ILP problem, solvable using an off-the-shelf solver and introduction of the computationally less demanding GRASP-based solver providing competitive solutions.

The remainder of the paper is organized as follows.
An overview of related work on power transmission line inspection and existing related formulations of the TSP is provided in the following section.
The addressed problem is defined in Section~\ref{sec:problem}.
The proposed ILP-based formulation is introduced in Section~\ref{sec:ilp} and the proposed heuristic GRASP-based solver in Section~\ref{sec:grasp}.
Results on optimal and heuristic solutions using a real-word test scenario are reported in Section~\ref{sec:results}.
Concluding remarks are detailed in Section~\ref{sec:conclusion}.

\section{Related Work}\label{sec:related}

The use of robotic vehicles in the PTL inspection is widespread, with possible solutions reviewed in surveys~\cite{katrasnik10survey,disyadej19smart}.
Multi-rotor UAVs in automated inspection utilizing path planning are evaluated in~\cite{he19multi}, where it is concluded that a high maneuvering precision makes them suitable for efficient vision-based inspection.
Various other used, but not herein discussed robots include line-hanging vehicle~\cite{lee13vision}, fixed-wing UAV~\cite{dong12capri}, Vertical Take-Off and Landing fixed-wing UAV~\cite{filho15vtol}, or UAV with wire landing capabilities~\cite{hamelin19line}.
Vision-based tracking of power-lines is evaluated in ~\cite{mendenez16vision} and \cite{zhou16tracking} reports on experimentally verified real-time automated tracking of power lines using a multi-rotor UAV.


The inspection planning can be formulated as a variant of the TSP~\cite{galceran13coverage},
\reviewii{%
e.g., applied in pylon components inspection using a Genetic Algorithm (GA)~\cite{tsp-in-ptl}.
The TSP-based formulation with a GA solution is proposed in~\cite{path-alg-uav} to address the PTL inspection with the maximum flight distance limit.
The authors of~\cite{2lrouting} propose the TSP-based Two-Layer Point-Arc Routing problem for the coordinated inspection using ground and aerial vehicles solved by ``Cluster First, Route Second'' and ``Route First, Split Second'' based heuristics.
}

\reviewiv{
The Set TSP, also called the Generalized TSP, has been addressed by heuristics~\cite{helsgaun2015gtsp} and ILP~\cite{laporte87stsp}.
The Cost constrained TSP~\cite{sokkappa90cost} is a suitable formulation for tasks where each target has associated value, and the problem is to visit targets with the maximal sum of the values within constrained tour cost, which is also known as the Orienteering Problem in the literature~\cite{penicka19ejor}.
A further generalization is addressed in~\cite{faigl19ral}; however, these approaches are for single tour problems.
}

Multi-route inspection can be formulated as a variant of the Multiple Traveling Salesmen Problem (MTSP)~\cite{bektas06omega} that can be solved using the ILP~\cite{miller60ilptsp}, which is known to be very computationally demanding, however.
Alternatively, less demanding GRASP heuristics have been employed in~\cite{guemri16or} for a problem similar to the herein addressed PTL problem.
The problem is referred to as the multi-product multi-vehicle inventory routing problem, which is a generalization of the inventory routing problem~\cite{bertazzi12jtl}.
Heuristics and integer programming formulations of the MTSP for UAV planning already exist, such as~\cite{Kitjacharoenchai2019} for UAV-based delivery in combination with trucks.
\reviewiii{%
Besides, soft computing techniques have been used for the MTSP, such as GA~\cite{mtsp-ga}, simulated annealing~\cite{mtsp-sa}, and neural networks~\cite{mtsp-ann,faigl16cin} to name a few.
}%
\reviewii{%
However, the proposed Multi-tour Set TSP formulation differs from the existing variants of the MTSP in the set generalization and travel budget.
}%
\reviewiii{%
Therefore, a novel solution is required, and the reader is referred to~\cite{cheikhrouhou21mtsp} for an overview of methods for the existing variants of the MTSP.
}

Based on the literature review and the best of the authors' knowledge, the problem of \reviewii{Multi-tour PTL wire} inspection with a limited travel budget has not been addressed by the existing approaches.
Therefore, we propose to formulate the problem using the ILP to find an optimal solution.
Furthermore, we address the computational challenges of the studied PTL inspection planning by the proposed heuristic inspired by the successful deployment of the GRASP-based approach to solving a similar problem in~\cite{guemri16or}.
GRASP is fine-tuned with the guidance of work discussing the specifics of tabu-search~\cite{tsub98tabu}, which is a part of the proposed novel meta-heuristic algorithm.

\section{Problem Statement}\label{sec:problem}
In the addressed PTL inspection, the goal is to traverse all given PTL segments in the shortest time possible.
The segments are defined as the power lines between two pylons, where each must be traversed in a single run to complete its inspection. 
However, the direction in which the segment is traversed is arbitrary.
The visit to each segment can be formulated as a vertex visit in a set of possible traverse directions, where every set should be visited exactly once.
It implies the problem can be formulated as a variant of the Generalized TSP~\cite{laportt83gtsp}, also called the Set TSP~\cite{manyam16uav}.
However, the formulation needs to include the practical limits of UAVs.

{
\begin{figure}[htbp]\centering
   \subfloat[
   Power-line segments between the pylons\label{fig:pylons}]{\includegraphics[width=.9\columnwidth]{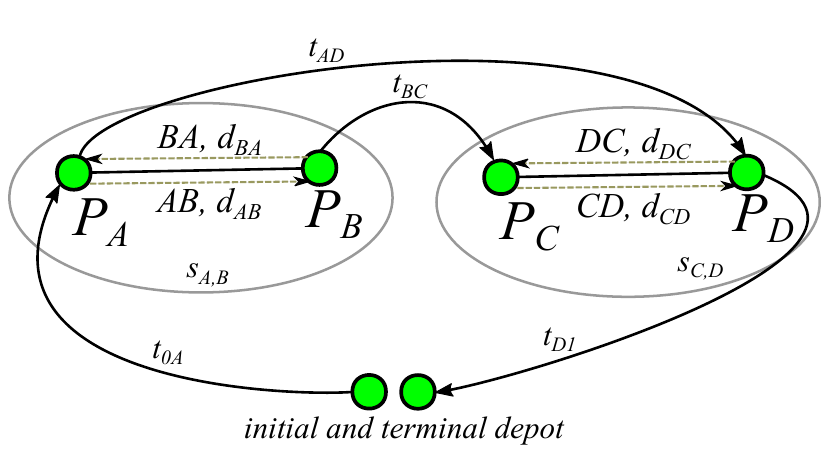}}

   \subfloat[Graph representation of the problem\label{fig:graph}]{\includegraphics[width=0.9\columnwidth]{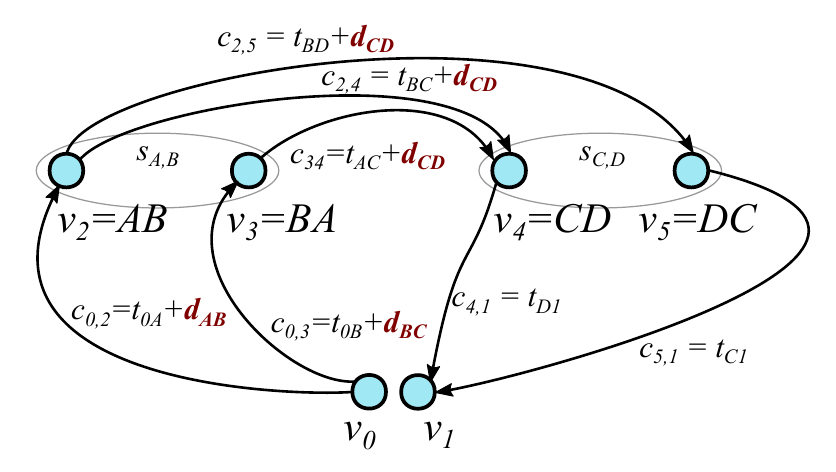}}
   \caption{Representation of the PTL inspection planning problem, where the pylons are denoted by alphabet letters for improved readability.
   (\textbf{a}) Each power line segment can be inspected in two possible directions, with the inspection speed being lower than overflight speed from one segment endpoint to another.
   (\textbf{b}) The vertices represent the power line segment visits in a specific direction; therefore, the edge cost is a sum of the travel cost from one segment to another and the cost of the target segment inspection.
   }
\end{figure}
}
\reviewii{First, the inspection plan must satisfy the UAV's limited flight time.
Additional constraints are the maximal flight speed and acceleration addressed in the estimation of flight times using a trajectory planner, making the planned routes physically feasible.
Multiple inspection tours are allowed as long as they begin and end at the same specific depot.
Based on the problem description, the notation used in the formal problem definition is as follows.}

\reviewii{We model the problem domain as a weighted oriented graph $G = (V, E,c)$ with a set of vertices $V$ representing visits to the PTL segments and a set of edges $E$.
A segment can be visited in an arbitrary direction, and therefore, the visits to the segments are partitioned into $n_s$ sets corresponding to $n_s$ physical segments. 
A segment $s_{A,B}=\{ v_{AB}, v_{BA}\}$ corresponds to two directions of visits to the segment between the pylons $P_A$ and $P_B$, see Fig.~\ref{fig:pylons}.
Each inspection tour is requested to start and terminate at the defined locations, and therefore $v_0 \in V$ and $v_1 \in V$ correspond to starting and termination depot, respectively.}
\reviewiv{An example valid route in Fig.~\ref{fig:pylons} is $v_0\rightarrow P_A \rightarrow P_B \rightarrow P_C \rightarrow P_D \rightarrow v_1$ in, which in the graph formulation Fig.~\ref{fig:graph} translates to $v_0\rightarrow v_2 \rightarrow v_4 \rightarrow v_1$.}

The graph $G$ contains $n=2+2n_s$ vertices $V=\{v_0,\ldots v_{2n_s},v_{2n_s+1}\}$ that are grouped into $2+n_s$ sets, where the first two sets contain vertices corresponding to the initial and termination locations.
Hence, the vertices can be referred to as integer values, and the edges can be referred to using integer indices of $V$, e.g., $e(v_2, v_5)=e_{2,5}$.
Furthermore, the first vertex index corresponding to the visit of a power line segment starts at $2$, and two directions of the possible segment visits are vertices $v_{2i}$ and $v_{2i+1}$.
Thus, $V=\{v_0,v_1\}\cup \bigcup_{i=1}^{n_s}S_i$, where the set $S_i=\{v_{2i},v_{2i+1}\}$ represents two possible visits of the $i$-th segment.

The particular flight times are determined by estimating flight trajectories connecting the segment endpoints and trajectories along with the segments.
The trajectories are subject to lateral flight dynamics constraints.
The constraints include the maximum flight velocity \Vmax, maximum inspection velocity \Vinsp, and acceleration \Amax.
The time of flight between the segment endpoints (pylons) $t$ is limited by the maximum UAV velocity \Vmax, and the time of the segment inspection $d$ is limited by the inspection velocity \Vinsp.
The edge cost $c(e_{i,j})$ (also referred as $c_{i,j}$) is the time spent on the travel to the segment endpoint from the endpoint of the previous segment added to the time spent on the inspecting the power line segment, except the edges to the terminal vertex $v_1$, as seen in Fig.~\ref{fig:graph}.
The practical feasibility that the UAV can track the used trajectories has been experimentally verified and the results are reported in Section~\ref{sec:results}.

Due to the limited travel budget \Tmax{}, all the power line segments cannot be visited in a single tour.
Therefore, we search for the optimal number of tours $\mathcal{T}=\{T_1,\ldots , T_{n_t}\}$ such that each tour $T_i\in \mathcal{T}$ originates at $v_{0}$, terminates at $v_1$, and each segment is visited exactly once in a union of all tours, i.e., only one of $v_{2i}$ and $v_{2i+1}$ is visited, because the PTL segment is represented by two vertices with the first vertex of the even index and the second incremented by one.
The travel cost of the tour $T_i\in \mathcal{T}$ has to satisfy the travel budget constraint $c(T)\le \Tmax$.

The PTL inspection planning problem can be formally defined as the \emph{Multi-tour Set Traveling Salesman Problem} (MS-TSP) depicted in Problem~\ref{problem:mt-set-tsp}.

\begin{problem}[Multi-tour Set Traveling Salesman Problem]\label{problem:mt-set-tsp}
   \begin{equation*}
      \begin{array}{lll}
	 \multicolumn{3}{l}{\displaystyle\operatorname{minimize}_{n_t,\mathcal{T}=\{T_1,\ldots,T_{n_t}\}}\, c(\mathcal{T})=\displaystyle\sum_{T\in \mathcal{T}}c(T)}\\
	 \multicolumn{2}{l}{~~\textrm{s.t.}} \\ 
	 \multicolumn{3}{l}{c(T)\le \Tmax \text{\rm~for }T\in \mathcal{T}
	 \text{and for each $S_i=\{v_{2i},v_{2i+1}\}$,}} \\
	 \multicolumn{3}{l}{1\le i \le n_s\text{, only $v_{2i}$ or exclusively $v_{2i+1}$ is visited by $\mathcal{T}$.}}
      \end{array}.
   \end{equation*}
\end{problem}

\section{ILP Formulation}\label{sec:ilp}
The optimal solution of the formulated MS-TSP can be obtained by a solution of the ILP problem formulation using the notation of the problem instance on a graph with $n$ vertices and precomputed individual costs between them as two-dimensional asymmetric cost matrix $\matr{C}\in\mathbb{R}^{n\times n}$, with the edge costs $c(e_{i,j})$, $0\le i,j \le n$.
The solution can be described by the three dimensional matrix of variables $\matr{X}\in\{0,1\}^{n_t\times n \times n}$, \reviewii{where the matrix element $x_{m,i,j}$ denotes the edge traversal from $i$ to $j$ by a tour $m$}, and $n_t$ is the maximal number of the tours.
The matrix $\matr{X}$ specifies which edges are used by the individual tours.
The tours are encoded in traversal matrix $\matr{T} \in \mathbb{N}^{n_t\times n}$, where elements $t_{m,i}$ denote the position of the vertex $i$ in the tour $m$. Thus, $\matr{T}$ specifies the order of the edges visits.

The initial and final locations are defined as vertices $v_0$ and $v_1$.
Furthermore, each segment is represented by two vertices $v_{2i}$ and $v_{2i+1}$ for $1\le i \le n_s$; therefore, the number of vertices $n$ is always even.
The ILP formulation is as follows.
\reviewii{
\begin{eqnarray}
  \mathcal{C} = \sum_{1\le m \le n_t} \sum_{1\le i \le n} \sum_{1 \le j \le n}c_{m,i,j}x_{m,i,j}\label{eq:objective}\\
    \sum_{i=2}^{n-1}x_{m,0,i} = 1,\text{ for } 0 \le m < n_t \label{eq:initial}\\
    \sum_{i=2}^{n-1}x_{m,i,1} = 1,\text{ for } 0 \le m < n_t \label{eq:final}\\
    \sum_{j=0}^{n-1}\sum_{m=0}^{n_t-1} (x_{m,2i,j} + x_{m,2i+1,j}) = 1, \text{ for } \ 1 \leq i < \frac{n}{2} \label{eq:pair1}\\
    \sum_{j=0}^{n-1}\sum_{m=0}^{n_t-1} (x_{m,j,2i} + x_{m,j,2i+1}) = 1, \text{ for } \ 1 \leq i < \frac{n}{2}\label{eq:pair2}\\
    \sum_{i=0}^{n-1} (x_{m,i,j} - x_{m,j,i}) = 0,\text{ for } 0\le m < n_t, 2 \leq j < n\label{eq:pair3}\\
    \sum_{(i,j)\in \matr{X}_m} c_{m,i,j}x_{m,i,j} \leq \Tmax,\text{ for } 0\le m < n_t \label{eq:budget}\\
    \begin{array}{r}
   t_{m,i} - t_{m,j} + x_{m,i,j} \leq n - 1,\text{ for } 2 \leq i \neq j \leq n,\\
    0 \le m < n_t\end{array} \label{eq:subtour}
\end{eqnarray}}

The objective \eqref{eq:objective} is to minimize the total solution cost~$\mathcal{C}$.
Constraints \eqref{eq:initial} and \eqref{eq:final} ensure that each tour starts at $v_0$ and terminates at $v_1$.
The requirement that each set (corresponding to two possible directions of how the power-line segment can be inspected) is entered and exited exactly once is ensured by \eqref{eq:pair1}--\eqref{eq:pair3}.
\reviewiii{%
   Constraints~\eqref{eq:pair1} state that for each pair of segment vertices $i$, the sum of all variables (representing the arrival to destination vertex $j$) over all routes $m$ is equal to 1.
Constraints \eqref{eq:pair3} ensure that for each tour $m$,  once a vertex $j$ is entered, the tour also exits it, where $i$ indexes all possible points of entry and exit for the vertex.
}
The travel budget \Tmax{} is constrained by~\eqref{eq:budget}, where $\matr{X}_m \in \{0,1\}^{n\times n}$ denotes the square sub-matrix of $\matr{X}$ containing variables for the tour $m$.
Finally, the sub-tour elimination constraint (Miller-Tucker-Zemlin~\cite{miller60ilptsp}) is included for each tour using elements of the matrix $\matr{T}$ in~\eqref{eq:subtour}.

The presented formulation can be used to solve the problem using off-the-shelf ILP solvers. The CPLEX~\cite{cplex2009v12} solver was used to obtain our baseline solution.

\section{Proposed Heuristic Solver}\label{sec:grasp}

Even though the solution of Problem~\ref{problem:mt-set-tsp} can be found for a particular value of $n_t$ by the presented ILP formulation, it requires high computational resources for finding the optimal solution that might not be practical when the inspection plan needs to be updated on-site in real-time.
It may happen that finding the optimal solution requires more time than performing an inspection with a suboptimal solution.
Thus, we propose a GRASP-based heuristic algorithm to find a feasible inspection plan quickly to follow the requirements of PTL operators from the AERIAL CORE consortium.

The proposed algorithm is based on the existing GRASP-based planner for the multi-product multi-vehicle routing problem~\cite{guemri16or} that we generalized to the Set TSP. 
In particular, the proposed planner employs two meta-heuristics.
The initial (but not necessarily feasible) solution is found by the \emph{Greedy Random search Procedure} (GRP) and is followed by the adaptive \emph{Tabu Search} (TS) to explore the solution neighborhood using a set of specific moves.
A soft constraint is used to limit the maximum tour cost during the search.
It is not guaranteed that the best solution found will be feasible with respect to~\Tmax.
Hence, multiple iterations with an increasing number of tours $n_t$ might be required. 

The soft constraint is violated if the tour cost $c(T)$ exceeds the maximum allowed cost \Tmax, as shown in Equation~\ref{eq:soft}. 
The constrained cost is referred to as $c_{con}(T)$.
\begin{equation}
\label{eq:soft}
    c_{con}(T) = 
    \begin{cases}
      c(T) & \text{if } c(T) \leq \Tmax,\\
      c(T) + (c(T)-\Tmax)k_c & \text{if } c(T) > \Tmax.
    \end{cases}
\end{equation}
The constant $k_c$ is chosen to be a sufficiently large number with regard to the tour costs.
Similarly to the ILP formulation, all costs between the vertices are pre-computed using the same trajectory planner to obtain feasible trajectories.
The real computational requirements of finding all trajectories are negligible compared to the combinatorial part of the problem.
The two parts of the GRASP-based solver are described in the following parts of this section.

\subsection{Greedy Random search Procedure (GRP)}
The initial solution $S_{initial}$ of is obtained using the GRP where the segments are iteratively placed into $S_{initial}$ from the set of unused segments $S_{available}$.
Possible insertions are stored in $I_{proposed}$, where the insertion $i_{s,m,p,d}\in I_{proposed}$ encodes a possible insertion of some segment $s$ into the tour $m$ at the position $p$ of the tour in the direction $d$.
\reviewi{Since the costs are pre-computed, the solution cost after insertion is known and denoted as Cost($i$) and is the cost $c_{con}(T)$ of the route $T$ that is being inserted into. 
It partially equalizes individual tour costs to reach a valid solution that does not violate \Tmax and leads to better solutions in the following TS.}

\begin{algorithm}[!b]
   \caption{Greedy Random search Procedure (GRP)}
   \label{alg:gr}
   \begin{algorithmic}[1] 
      \State $S_{initial}\gets \emptyset$;
      \State $S_{available} \gets S$;
      \While{ size($S_{available}$) $\neq 0$}
      \State Fill $I_{proposed}$ with all possible insertions;
      \State Order elements of $I_{proposed}$ by Cost($i$);
      \State Fill $I_{RCL}$ with a subset of $I_{proposed}$;
      \State $i_{selected} \gets$ Random $i$ from $I_{RCL}$;
      \State Insert $i_{selected}$ into the route $m$ and update $S_{initial}$;
      \State Remove $i_{selected}$, and other insertions referring to same segment set from $S_{available}$;
      \EndWhile
      \State \Return $S_{initial}$;
   \end{algorithmic}
\end{algorithm}
The Boolean variable indicates the inspection direction $d$, however it can be an integer value in the case of further generalization.
The \emph{Restricted Candidate List} (RCL) of the best insertion candidates denoted $I_{RCL}$ is utilized in the randomized search of the GRP.
The $I_{RCL}$ is empirically chosen to contain $\SI{25}{\percent}$ of all proposed insertions to get varied initial solutions.
The GRP is summarized in Algorithm~\ref{alg:gr}.

\subsection{Adaptive Tabu Search}
After the GRP determines the initial solution, the adaptive TS is applied over several iterations to the routes.
The solution space is explored by repeatedly selecting from four distinct moves using the weighted roulette wheel~\cite{guemri16or}. 
Similar moves are used to specify the solution neighborhood with one move for direction-switching added, which hastens the convergence to a lower solution cost.
\begin{itemize}
   \item Move 1 - 
      \textit{Random-shift}: Remove a random segment from a random route and insert it into a random position in the same or other route. 
      The move adds randomness to the search, increasing its weight in every iteration.
   \item Move 2 - 
      \textit{Best-shift}: Remove a random segment from the route and insert it into the best position in the same or other route.
      The cheapest (best) solution is determined by the exhaustive search from all possible insertions. 
   \item Move 3 - 
      \textit{Best-swap}: Swap a random segment from a random route with the best segment in the same or other route. 
      The best swap is, like in the previous move, found through an exhaustive search from all swaps.
   \item Move 4 - 
      \textit{Best-direction-switch}: Switch a traversal direction for the best segment that is also evaluated over all segments.
\end{itemize}
Both directions of segments are evaluated for the first three moves and the best is/are chosen.

The score weight of each move ($w_i, i \in 1 \ldots 4$) is initialized to the value $w_0$ and is increased every iteration according to the performance of moves.
Two prize values are used in the employed adaptive TS: $p1$ and $p2$.
The price $p1$ is added to the best move in the neighborhood, and $p2$ is added to the move that improved the best overall TS solution.
The value of $p1$ is added to the weight $w_1$ every iteration to promote randomness in neighborhood exploration.
For every $R_t$ iterations, the score weights are reinitialized to the value $w_0$.
The neighborhood of the size $n^2$ is explored using the tabu list $T_{list}$ of the size $n/4$. 

\begin{algorithm}[htbp]
   \caption{Adaptive Tabu Search}\label{alg:tabu} 
   \begin{algorithmic}[1] 
      \Require{Some existing solution $S$.}
      \State $S_{best} \gets S$;
      \State $S_{current} \gets S$;
      \State Insert $S_{current}$ into $T_{list}$;
      \While{Stopping criterion is not met};
      \State Create a solution neighborhood $N(S_{current})$;
      \State $S_{current} \gets $Best\{$S \in N(S_{current})$ and $S \notin T_{list}$\};
      \If{Cost($S_{current}$) $<$ Cost($S_{best}$)}
      \State $S_{best} \gets S_{current}$;
      \EndIf
      \EndWhile
      \State \Return $S_{best}$;
   \end{algorithmic}
\end{algorithm}
The TS procedure is outlined in Algorithm~\ref{alg:tabu}, where $S$ denotes some existing solution, $T_{list}$ is the tabu list of explored solutions, and $N(S)$ is the neighborhood of some solution $S$.
\reviewi{The stopping criterion in Algorithm~\ref{alg:tabu} is when there is no improvement of $S_{best}$ in the last 50 iterations of creating a solution neighborhood. 
This empirically selected criterion is chosen to acquire the best possible solution to be compared with the CPLEX solution. 
Specific application of the GRASP procedure might prefer a different criterion, such as a time limit in real-time operation.}

\section{Results}\label{sec:results}

\begin{figure}[!htb]\centering
   \includegraphics[width=0.9\columnwidth]{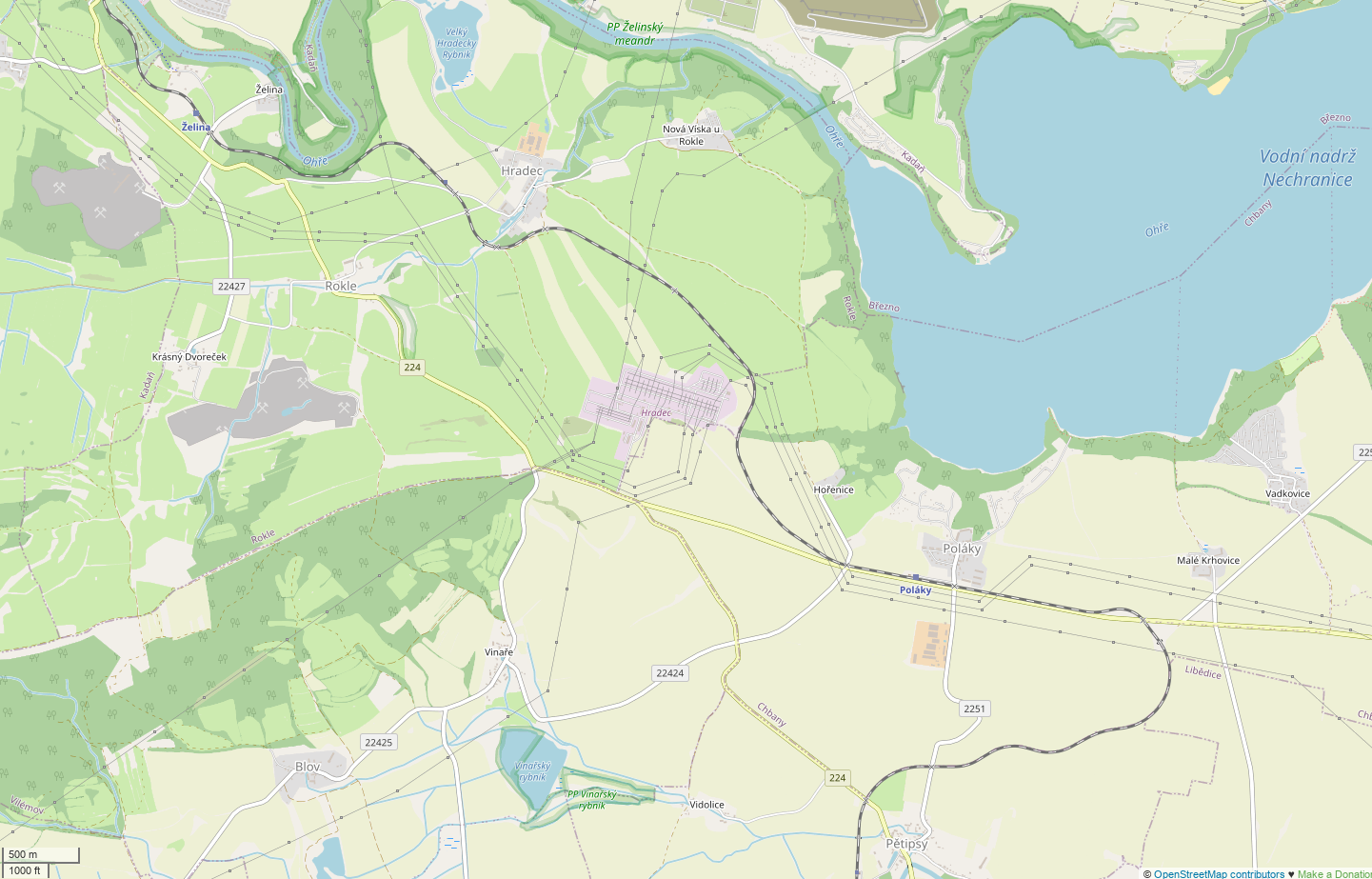}
   \caption{Map of Nechranice substation and its surroundings used to create realistic PTL inspection planning problem instances. The map was acquired from Open Street Map~\cite{OpenStreetMap} database.}\label{fig:map}
\end{figure}

The feasibility of the proposed PTL inspection planning has been empirically validated on a real-world dataset~\cite{mrs-mstsp} of the PTLs originating from the substation Nechranice, courtesy of ČEPS, a.s. (Czech power transmission infrastructure institution, a member of the AERIAL~CORE advisory board).
A~map showing an overview of the PTL layout surrounding the substation is depicted in Fig.~\ref{fig:map}.

The S-JTSK coordinates of pylons were transformed into the Cartesian system with the origin at the substation.
The benchmark problem instances were sampled from the area around the Nechranice substation using the perimeter~\Dmax, starting with a small number of segments that increases with \Dmax{}, as indicated in Table~\ref{table_dist}.
\begin{table}[h]\centering
   \caption{Size of the Benchmark Instances}\label{table_dist}
   \resizebox{\columnwidth}{!}{\begin{tabular}{l  r r r r r r r r r}
   \noalign{\hrule height 1.1pt}\noalign{\smallskip}
   \Dmax [\si{m}] & 500 & 600 & 700 & 800 & 900 & 1\,000 & 1\,200 & 2\,000 & 5\,000\\
   $n_{s}$ & 15 & 23 & 28 & 33 & 40 & 43 & 51 & 76 & 172\\
   \noalign{\hrule height 1.1pt}
\end{tabular}
}
\end{table}
Furthermore, several limits on the flight time \Tmax{} have been chosen for each particular scenario to evaluate the influence of the inspection area defined by \Dmax{} and the maximal flight time per a single tour.
The travel budget \Tmax{} is set proportionally to \Dmax{} to obtain feasible solutions for $n_t\in\{3,\ldots,6\}$ that corresponds to tens of minutes of flight time for \Dmax{} in hundreds of meters and reaches up to several hours for finding a feasible solution of the largest instances.

The travel costs have been determined by estimating flight trajectories of a real UAV constrained by its flight dynamics. 
The cost of travel between the segments is limited by the maximum flight velocity $\Vmax = \SI{5}{\meter\per\second}$ and the traversal cost during the segment inspection is limited by the maximum inspection velocity $\Vinsp = \SI{1}{\meter\per\second}$.
The maximum acceleration is limited to $\Amax = \SI{2.5}{\meter\per\square\second}$.
Experimental validation of the costs is reported in Section~\ref{sec:experiment}.

Each particular instance is solved optimally using the proposed ILP formulation for the lowest $n_t$, for which the ILP solver provides the solution within the maximal computational time of up to \SI{24}{\hour}.
Note that the ILP solver finds the optimal solution for the increasing number of tours, which is usually less demanding. However, such a solution is not the optimal solution of the original problem with respect to the minimal number of tours as stated in Problem~\ref{problem:mt-set-tsp}.

Both proposed methods, the ILP-based and GRASP-based approaches, have been implemented in C++.
The ILP is solved using CPLEX 12.8.0~\cite{cplex2009v12} implemented in C++ using the ILOG Concert technology library.
Computations were performed on the Intel Xeon processors (2.2--3.3\si{\giga\hertz}) limited to utilization of a single core.
Each CPLEX instance has been limited by \SI{100}{\giga\byte} of RAM.
Without a prior initial solution, the computational requirements of solving the ILP are exceptionally high, and most of the herein reported results would not be computed in a limited time.
Therefore, the CPLEX solver has been initialized a solution found by the proposed GRASP-based solver, which computational cost is negligible compared to the cost of the optimal solution.
The GRASP-based solver has been parameterized as follows.

The candidate list size is \SI{25}{\percent} of the possible insertion list size.
The initial move weight in the TS is $w_0 = 5$, the prize weights $p_1 = 1$, $p_2 = 5$, and number of iterations after which the weights are reset is set to $w_r = 5$.
As outlined in~\cite{tsub98tabu}, the size of tabu lists should be kept small. We therefore opted for $n/4$, where $n$ is the number of segments manipulated.
The size of the solution neighborhood has been fixed at $n$.
The stopping criterion is 50 iterations without improvement of the best solution.
Because the GRASP is stochastic and relatively inexpensive to compute, each instance has been solved 30 times.
Both the best solution found and mean solution values among the trials were reported.

\subsection{Computational Evaluation}
The performance of both proposed solvers \reviewv{including a basic Greedy Random (GR) search} has been studied for the size of the instances defined by \Dmax{}, the travel budget \Tmax{}, and the number of inspection tours $n_t$.
The evaluation results are depicted in Table~\ref{tbl:results}, where the lowest optimal solution costs are in bold.
\reviewv{The best solution cost provided by the GR algorithm is also included in case it managed to provide a valid solution among the performed trials.
The $\%SR$ denotes success rate of this search.}
In all examined benchmark instances, the best solutions are found by the ILP problem formulation for the lowest possible $n_t$ that is considered the optimal solution of the instance with the particular cost $\mathcal{C}^*$. 
Therefore, we measure the quality of the ILP-based solution for the increased number of tours $n_t$, and also for the heuristic GRASP-based solutions as the percentage deviation of the solution cost to the optimal solution cost $\%\textrm{PDB}=({\mathcal{C}-\mathcal{C}^*})/{\mathcal{C}^*} \cdot 100\%$.
The computational time of the optimal ILP-based solution and the solution of all 30 trials by the GRASP is denoted t and reported in seconds.
Thus, the times can be compared with the solution cost $\mathcal{C}$ that is reported as the required total flight time in seconds.
Additionally for the GRASP, the mean computational time t\textsubscript{mean} per a single trial is reported together with the mean solution quality $\mathcal{C}_\text{mean}$, which is reported as the percentage deviation of the mean solution cost to the optimal solution cost $\%\textrm{PDM}=(\mathcal{C}_{\text{mean}}-\mathcal{C}^*)/\mathcal{C}^*$ that measures the robustness of the solution among the performed trials.

\begin{figure}[tb]\centering
   \includegraphics[width=0.99\columnwidth,clip,]{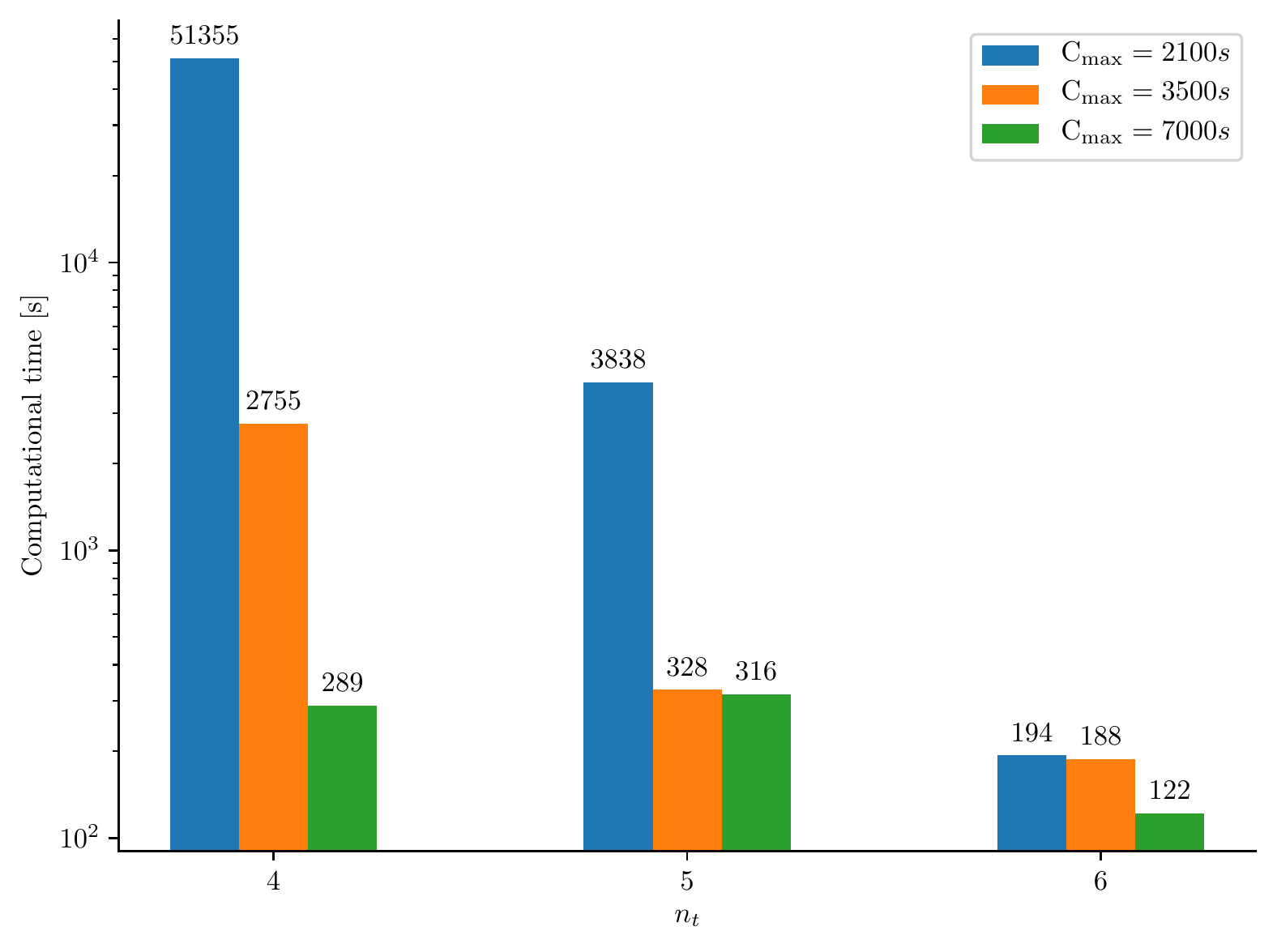}
   \caption{Computational requirements to solve the ILP formulation for instances with $\Dmax = \SI{700}{\meter}$, travel budget \Cmax{}, and $n_t$ tours. The $y-$axis is in log-scale. Solution costs are above their respective bars.\label{fig:d700}
   }
\end{figure}

\begin{table*}[!htb]\centering
   \caption{Computational Results}\label{tbl:results}
   \vspace{-1em}
   \scalebox{0.95}{
      \resizebox{1\textwidth}{!}{
	 \begin{tabular}{
	       r
	       r
	       c
	       r
	       r
	       r
	       r
	       r
	       r
	       r
	       r
	       r
	       r
         r
         r
         r
         r
         r
	    }
	    \noalign{\hrule height 1.1pt}\noalign{\smallskip}
	    \multicolumn{1}{c}{\Dmax} &\multicolumn{1}{c}{\Cmax} & \multirow{2}{*}{$n_{t}$}
      &\multicolumn{3}{c}{ILP} && \multicolumn{6}{c}{GRASP}&&\multicolumn{4}{c}{GR$^\dagger$} \\
      \cline{4-6}\cline{8-13}\cline{15-18}
	    \noalign{\vspace{0.25em}}
	    \multicolumn{1}{c}{[\si{m}]} & \multicolumn{1}{c}{[\si{s}]} &
	    & \multicolumn{1}{c}{t [\si{s}]} & \multicolumn{1}{c}{$\mathcal{C}$ [\si{s}]} & \multicolumn{1}{c}{\scriptsize\%PDB} 
      && \multicolumn{1}{c}{t\textsubscript{mean} [\si{s}]} & \multicolumn{1}{c}{t [\si{s}]} & \multicolumn{1}{c}{$\mathcal{C}_{\text{mean}}$ [\si{s}]} & \multicolumn{1}{c}{\scriptsize\%PDM}& \multicolumn{1}{c}{$\mathcal{C}$ [\si{s}]}& \multicolumn{1}{c}{\scriptsize{\%PDB}}&&\multicolumn{1}{c}{t [\si{s}]} &\multicolumn{1}{c}{$\mathcal{C}$ [\si{s}]}& \multicolumn{1}{c}{\scriptsize{\%PDB}} & \multicolumn{1}{c}{\scriptsize{\%SR}}
	    \\
	    \noalign{\vspace{0.1em}} \noalign{\hrule height 1.1pt}\noalign{\smallskip}
	    
{500}&{1\,000}&{4}&{7.4}&\textbf{3\,178.6}&{0.0}&&{0.05}&{1.64}&{3\,506.5}&{10.3}&{3\,499.8}&{10.1}&&{0.05}& - &- &{0}\\ 
{500}&{1\,000}&{5}&{2.6}&3\,221.4&{1.3}&&{0.08}&{2.27}&{3\,397.4}&{6.9}&{3\,371.0}&{6.0}&&{0.06}&-&-&{0}\\ 
{500}&{1\,000}&{6}&{0.9}&3\,328.4&{4.7}&&{0.04}&{1.34}&{3\,566.6}&{12.2}&{3\,478.4}&{9.4}&&{0.06}&-&-&{0}\\ 
\hline\noalign{\vspace{0.25em}}
{500}&{1\,500}&{3}&{1.9}&\textbf{3\,053.2}&{0.0}&&{0.06}&{1.77}&{3\,281.7}&{7.5}&{3\,262.4}&{6.9}&&{0.04}&{-}&-{}&{0}\\ 
{500}&{1\,500}&{4}&{1.4}&3\,087.9&{1.1}&&{0.12}&{3.47}&{3\,321.7}&{8.8}&{3\,260.1}&{6.8}&&{0.04}&{5\,228.8}&{71.3}&{62}\\ 
{500}&{1\,500}&{5}&{1.2}&3\,194.9&{4.6}&&{0.05}&{1.49}&{3\,458.4}&{13.3}&{3\,448.1}&{12.9}&&{0.05}&{5\,063.1}&{65.8}&{100}\\ 
{500}&{1\,500}&{6}&{0.8}&3\,296.1&{8.0}&&{0.05}&{1.58}&{3\,492.2}&{14.4}&{3\,452.7}&{13.1}&&{0.05}&{5\,637.1}&{84.6}&{100}\\ 
\hline\noalign{\vspace{0.25em}}
{600}&{1\,800}&{3}&{74.1}&\textbf{5\,143.5}&{0.0}&&{0.07}&{2.15}&{5\,582.9}&{8.5}&{5\,497.5}&{6.8}&&{0.13}&{-}&{-}&{0}\\ 
{600}&{1\,800}&{4}&{46.3}&5\,176.8&{0.7}&&{0.15}&{4.63}&{5\,645.5}&{9.8}&{5\,590.7}&{8.7}&&{0.14}&{-}&{-}&{0}\\
{600}&{1\,800}&{5}&{35.5}&5\,244.6&{2.0}&&{0.11}&{3.25}&{5\,764.9}&{12.1}&{5\,705.2}&{10.9}&&{0.15}&{8\,608.0}&{67.4}&{19}\\ 
{600}&{1\,800}&{6}&{39.4}&5\,356.3&{4.1}&&{0.13}&{4.01}&{5\,843.6}&{13.6}&{5\,701.6}&{10.9}&&{0.18}&{8\,501.0}&{65.3}&{81}\\ 
\hline\noalign{\vspace{0.25em}}
{600}&{3\,000}&{3}&{12.1}&\textbf{5\,090.4}&{0.0}&&{0.18}&{5.36}&{5\,562.5}&{9.3}&{5\,460.4}&{7.3}&&{0.13}&{8\,473.0}&{66.5}&{57}\\ 
{600}&{3\,000}&{4}&{11.9}&5\,162.0&{1.4}&&{0.08}&{2.49}&{5\,595.4}&{9.9}&{5\,464.2}&{7.3}&&{0.14}&{8\,750.1}&{71.9}&{100}\\ 
{600}&{3\,000}&{5}&{10.9}&5\,229.8&{2.7}&&{0.13}&{3.78}&{5\,698.6}&{11.9}&{5\,661.9}&{11.2}&&{0.16}&{9\,098.5}&{78.7}&{100}\\ 
{600}&{3\,000}&{6}&{12.9}&5\,344.1&{5.0}&&{0.16}&{4.84}&{5\,878.6}&{15.5}&{5\,796.4}&{13.9}&&{0.16}&{8\,994.1}&{76.7}&{100}\\ 
\hline\noalign{\vspace{0.25em}}
{700}&{2\,100}&{4}&{51\,355.1}&\textbf{6\,268.3}&{0.0}&&{0.17}&{5.21}&{6\,951.6}&{10.9}&{6\,736.5}&{7.5}&&{0.26}&{-}&{-}&{0}\\ 
{700}&{2\,100}&{5}&{3\,838.8}&6\,321.1&{0.8}&&{0.14}&{4.08}&{6\,986.3}&{11.5}&{6\,717.4}&{7.2}&&{0.26}&{-}&{-}&{0}\\ 
{700}&{2\,100}&{6}&{194.3}&6\,416.9&{2.4}&&{0.20}&{6.13}&{7\,042.3}&{12.3}&{6\,928.9}&{10.5}&&{0.27}&{11\,247.0}&{79.4}&{100}\\ 
\hline\noalign{\vspace{0.25em}}
{700}&{3\,500}&{3}&{6\,401.6}&\textbf{6\,161.8}&{0.0}&&{0.29}&{8.61}&{6\,700.0}&{8.7}&{6\,526.1}&{5.9}&&{0.24}&{-}&{-}&{0}\\ 
{700}&{3\,500}&{4}&{2\,755.7}&6\,222.4&{1.0}&&{0.19}&{5.71}&{6\,748.6}&{9.5}&{6\,637.1}&{7.7}&&{0.25}&{10\,744.4}&{74.4}&{95}\\ 
{700}&{3\,500}&{5}&{328.7}&6\,289.7&{2.1}&&{0.43}&{12.88}&{6\,880.6}&{11.7}&{6\,716.0}&{9.0}&&{0.26}&{11\,434.3}&{85.6}&{100}\\ 
{700}&{3\,500}&{6}&{188.4}&6\,415.7&{4.1}&&{0.20}&{6.01}&{7\,118.6}&{15.5}&{7\,024.6}&{14.0}&&{0.28}&{10\,676.7}&{73.3}&{100}\\ 
\hline\noalign{\vspace{0.25em}}
{800}&{2\,400}&{5}&{15\,765.0}&\textbf{7\,510.0}&{0.0}&&{0.19}&{5.77}&{8\,420.8}&{12.1}&{8\,234.7}&{9.6}&&{0.40}&{-}&{-}&{0}\\ 
{800}&{2\,400}&{6}&{2\,807.3}&7\,602.4&{1.2}&&{0.28}&{8.54}&{8\,421.3}&{12.1}&{8\,288.1}&{10.4}&&{0.42}&{-}&{-}&{0}\\ 
\hline\noalign{\vspace{0.25em}}
{800}&{4\,000}&{3}&{24\,120.2}&\textbf{7\,369.2}&{0.0}&&{0.39}&{11.64}&{8\,058.1}&{9.3}&{7\,879.4}&{6.9}&&{0.38}&{-}&{-}&{0}\\ 
{800}&{4\,000}&{4}&{769.0}&7\,404.0&{0.5}&&{0.35}&{10.56}&{8\,210.2}&{11.4}&{8\,020.5}&{8.8}&&{0.39}&{13\,000.5}&{76.4}&{100}\\ 
{800}&{4\,000}&{5}&{24\,379.7}&7\,488.0&{1.6}&&{0.46}&{13.91}&{8\,285.1}&{12.4}&{8\,081.8}&{9.7}&&{0.40}&{12\,807.7}&{73.8}&{100}\\ 
{800}&{4\,000}&{6}&{1\,770.0}&7\,602.4&{3.2}&&{0.49}&{14.73}&{8\,482.1}&{15.1}&{8\,289.6}&{12.5}&&{0.41}&{13\,613.2}&{84.7}&{100}\\ 
\hline\noalign{\vspace{0.25em}}
{900}&{2\,700}&{6}&{10\,661.1}&\textbf{9\,014.4}&{0.0}&&{0.59}&{17.77}&{10\,210.4}&{13.3}&{9\,933.2}&{10.2}&&{0.67}&{-}&{-}&{0}\\ 
\hline\noalign{\vspace{0.25em}}
{900}&{4\,500}&{3}&{35\,372.2}&\textbf{8\,782.4}&{0.0}&&{0.64}&{19.05}&{9\,804.2}&{11.6}&{9\,600.3}&{9.3}&&{0.63}&{-}&{-}&{0}\\ 
{900}&{4\,500}&{4}&{1\,004.2}&8\,817.1&{0.4}&&{0.53}&{15.87}&{9\,906.9}&{12.8}&{9\,631.8}&{9.7}&&{0.64}&{15\,503.3}&{76.5}&{100}\\ 
{900}&{4\,500}&{5}&{26\,029.7}&8\,884.4&{1.2}&&{0.87}&{25.97}&{10\,108.4}&{15.1}&{9\,809.0}&{11.7}&&{0.64}&{15\,896.3}&{81.0}&{100}\\ 
{900}&{4\,500}&{6}&{3\,506.5}&9\,006.6&{2.6}&&{0.52}&{15.47}&{10\,246.7}&{16.7}&{9\,986.0}&{13.7}&&{0.67}&{16\,090.8}&{83.2}&{100}\\ 
\hline\noalign{\vspace{0.25em}}
{1000}&{3\,000}&{6}&{21\,174.5}&\textbf{9\,887.6}&{0.0}&&{0.65}&{19.51}&{11\,130.4}&{12.6}&{10\,870.0}&{9.9}&&{0.80}&{-}&{-}&{0}\\ 
\hline\noalign{\vspace{0.25em}}
{1000}&{5\,000}&{4}&{86\,370.0}&\textbf{9\,712.7}&{0.0}&&{0.62}&{18.48}&{10\,845.0}&{11.7}&{10\,544.2}&{8.60}&&{0.76}&{16\,970.2}&{74.7}&{100}\\ 
{1000}&{5\,000}&{6}&{13\,605.2}&9\,887.6&{1.8}&&{0.63}&{18.97}&{11\,112.0}&{14.4}&{10\,917.7}&{12.4}&&{0.82}&{17\,442.7}&{79.6}&{100}\\ 
\hline\noalign{\vspace{0.25em}}
{1100}&{5\,500}&{4}&{86\,370.1}&\textbf{11\,113.4}&{0.0}&&{1.50}&{45.14}&{12\,407.4}&{11.6}&{12\,144.0}&{9.3}&&{1.02}&{19\,603.2}&{76.4}&{100}\\ 
\hline\noalign{\vspace{0.25em}}
{1100}&{11\,000}&{3}&{59\,362.5}&\textbf{10\,995.0}&{0.0}&&{1.11}&{33.38}&{12\,196.8}&{10.9}&{11\,988.4}&{9.0}&&{0.98}&{17\,879.1}&{62.6}&{100}\\ 
{1100}&{11\,000}&{4}&{86\,370.0}&11\,078.4&{0.8}&&{0.68}&{20.27}&{12\,372.6}&{12.5}&{12\,168.8}&{10.7}&&{0.98}&{18\,963.6}&{72.5}&{100}\\ 
{1100}&{11\,000}&{5}&{9\,371.1}&11\,179.8&{1.7}&&{0.77}&{23.21}&{12\,565.2}&{14.3}&{12\,233.3}&{11.3}&&{0.97}&{19\,021.9}&{73.0}&{100}\\ 
\hline\noalign{\vspace{0.25em}}
{1200}&{12\,000}&{4}&{86\,370.1}&\textbf{12\,453.0}&{0.0}&&{0.76}&{22.80}&{13\,870.3}&{11.4}&{13\,567.6}&{9.0}&&{1.25}&{21\,502.8}&{72.7}&{100}\\ 
{1200}&{12\,000}&{5}&{86\,370.0}&12\,564.0&{0.9}&&{1.03}&{30.83}&{14\,044.0}&{12.8}&{13\,502.2}&{8.4}&&{1.23}&{22\,327.5}&{79.3}&{100}\\ 
\hline\noalign{\vspace{0.25em}}
{1300}&{13\,000}&{3}&{86\,370.0}&\textbf{13\,143.8}&{0.0}&&{1.31}&{39.33}&{14\,476.4}&{10.1}&{14\,156.6}&{7.7}&&{1.32}&{22\,278.2}&{69.5}&{100}\\ 
\hline\noalign{\vspace{0.25em}}
{1500}&{15\,000}&{4}&{86\,370.1}&\textbf{15\,860.6}&{0.0}&&{1.75}&{9.81}&{17\,637.8}&{11.2}&{17\,258.8}&{8.8}&&{2.03}&{26\,821.4}&{69.1}&{100}\\

	    \noalign{\hrule height 1.1pt}\noalign{\vskip0.25em}
	    \multicolumn{12}{l}{$^\dagger$\footnotesize Missing values indicate no feasible solution has been found among the performed trials for a particular problem instance.}
	 \end{tabular}
	 }
	 }
	 \vspace{-1em}
\end{table*}

\begin{figure}[!htb]\centering
   \includegraphics[width=0.9\columnwidth]{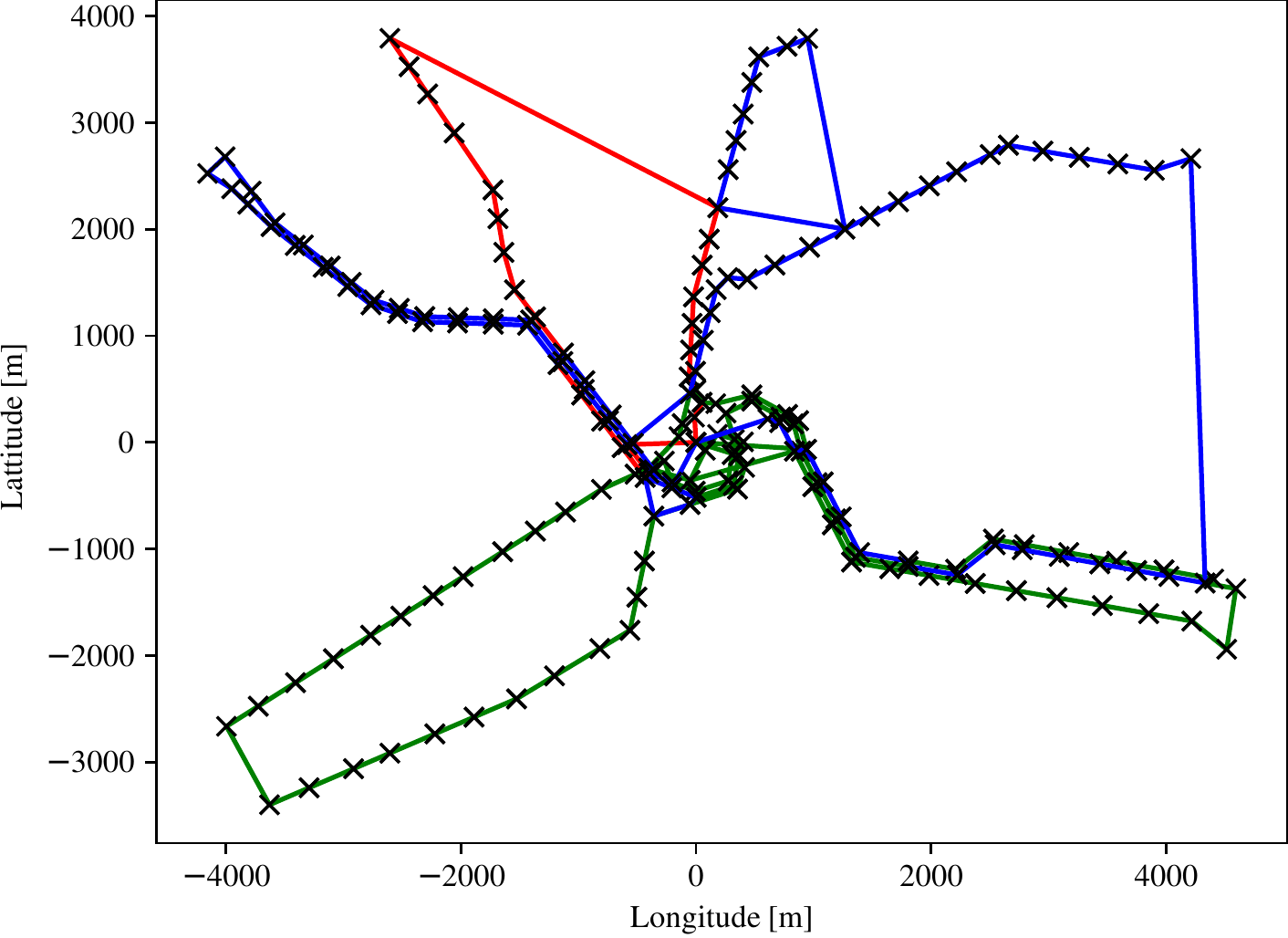}
   \caption{Example of the solution found by the GRASP for the largest examined instance for $\Dmax=\SI{5}{\kilo\meter}$ with $n_s=172$ segments and $n_t=3$.\label{fig:sol_big}
   }
\end{figure}

The result indicates that the ILP is very demanding.
Moreover, the computational time spikes unpredictably depending on the mixture of constraints and quality of the initial solution.
Generally, it tends to increase exponentially with tighter \Tmax{} and decreasing $n_t$, as is further shown in Fig.~\ref{fig:d700}.

The computational cost of the proposed heuristic GRASP-based solver is significantly lower than the ILP, and scales with the size of the instance defined by \Dmax{} and tight constraints on \Tmax{} and $n_t$.
Solutions are provided in about \SI{1}{\minute} for the largest instance depicted in Fig.~\ref{fig:sol_big}.
In the presented results for 30 trials, the solution cost variance is about $\SI{17}{\percent}$ and \%PDM is relatively high, but not exceeding \SI{20}{\percent}, which indicates a high variance in the found solutions.
The measured \%PDB of GRASP is not exceeding \SI{15}{\percent}.

\subsubsection*{Discussion} - 
Based on the reported results, the trade-off between almost instantaneously solving the ILP to improve the initial solution provided by GRASP is mostly unfavorable in the case of limited computational time.
The benefit of the optimal solution rapidly decreases with tighter constraints or larger instances.
Thus, it might be more suitable to deploy heuristic solutions, which are about tens or even hundreds of seconds longer than the optimal solutions that require significantly longer computational times.
On the other hand, the results also indicate that with the increasing number of tours~$n_t$, the ILP-based solution might still be better than a heuristic solution for a lower number of tours with significantly lower computational time to find the optimal solution.
Here, it is worth mentioning that the lowest feasible $n_t$ is not necessarily optimal for \Tmax.
The coincidence of the lowest $n_t$ in the presented results arises from the problem geometry of closely spaced star-shaped PTLs and the flight cost between the segment inspections being cheaper than the inspection cost.
The optimal value of $n_t$ depends on the use-case and which optimality criterion is preferred, i.e., the lowest number of tours or the lowest overall flight time.

\subsection{Experimental Validation}\label{sec:experiment}

\begin{figure}[t]
   \subfloat[Experimental UAV above line\label{fig:uav-above-line}]{\includegraphics[width=.45\columnwidth]{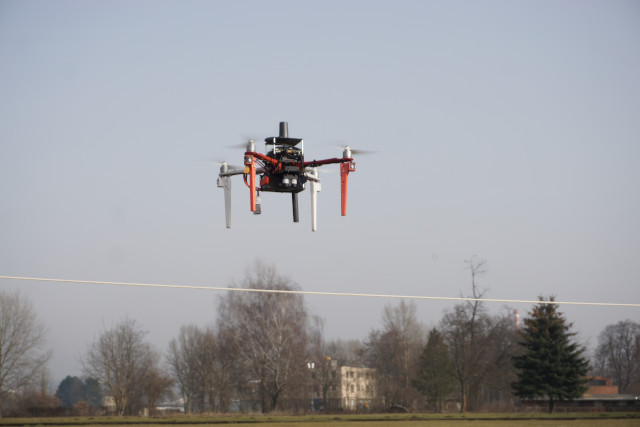}}\hfill
   \subfloat[Trajectory shown in rviz\label{fig:trajectory-rviz}]{\includegraphics[width=.49\columnwidth]{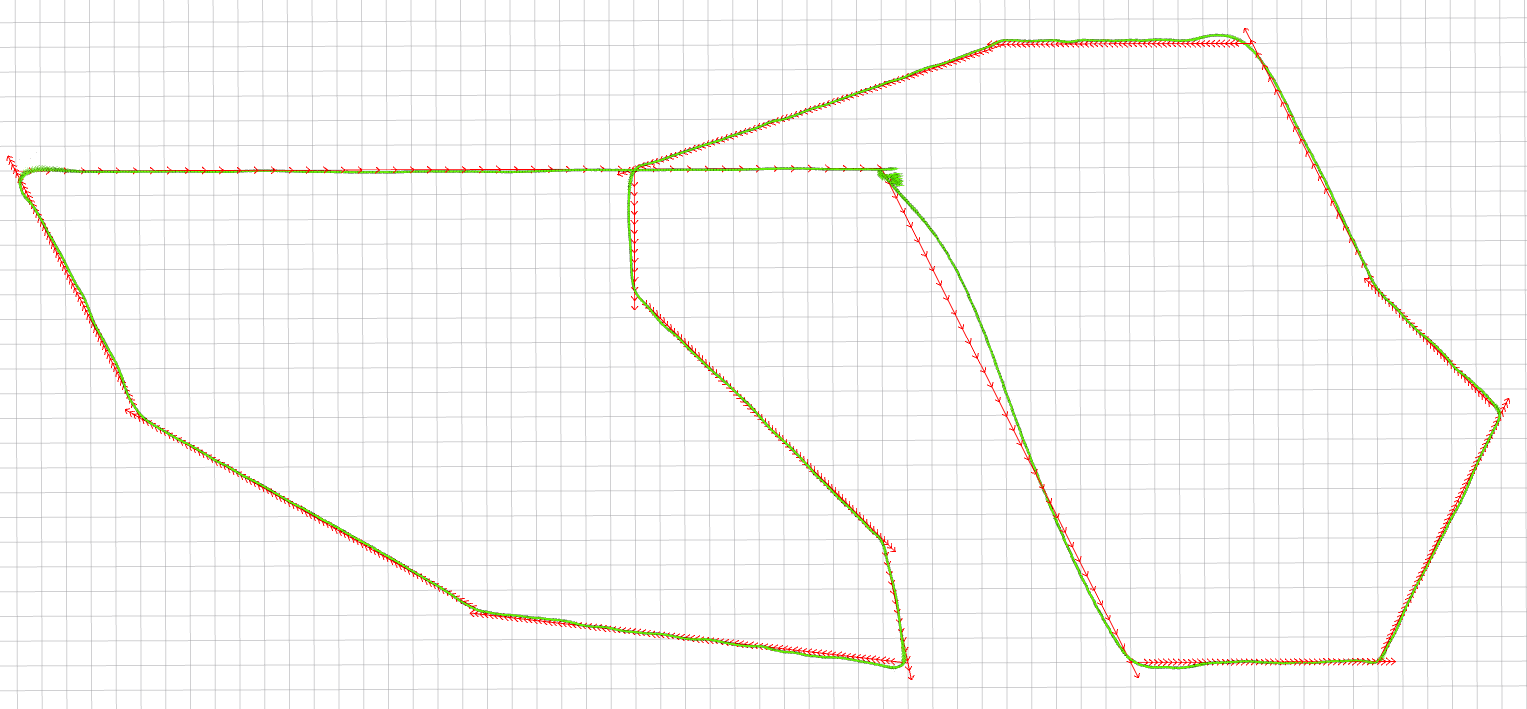}}
   \caption{The multi-rotor UAV used during the experiment \textbf{(a)} was equipped with Reach M2 GNSS module for \emph{Real-Time Kinematic} (RTK) positioning. Time sampled trajectory ($\Delta t = \SI{0.2}{\second}$) over segments \textbf{(b)} (shown in red) was tracked with regard to the UAV constraints (shown in green). Grid resolution is \SI{1}{\meter}. Video from the experiment is available at~\cite{mrs-mstsp}.}
\end{figure}

The practical feasibility of the proposed PTL inspection planning approach has been experimentally verified with real UAV in a testing polygon.
The experimental validation has been carried out using the control architecture~\cite{baca2018model} and UAV platforms designed for the AERIAL~CORE project~\cite{silano2021ral}.
A multi-rotor UAV used during the inspection is depicted in Fig.~\ref{fig:uav-above-line} and a trajectory containing a mix of segment lines is shown in Fig.~\ref{fig:trajectory-rviz}.
The experimental deployment supports the feasibility of the presented approach and verifies the created benchmark instances as realistic.
\reviewiii{Besides, it also supports feasibility of the estimated trajectory costs found by the trajectory planner with the employed UAV controller.}

\section{Conclusion}\label{sec:conclusion}
The PTL inspection planning with aerial vehicles has been addressed by the novel generalization of the TSP called the Multi-tour Set TSP.
The optimal ILP-based formulation is proposed together with a less computationally demanding GRASP-based heuristic approach.
Both solvers have been empirically evaluated and the feasibility of the proposed approach has been experimentally verified with real vehicles.
Additionally, a realistic dataset has been created and made available for benchmarking the proposed approach and for eventual further approaches.
The main benefit of the presented GRASP-based solver is in low computational requirements.
It can provide solutions on-site almost instantaneously, where a change of the inspection plan may be needed as required by the PTL operators and end-users of the proposed approach.
The possible future work lies with improving the scalability for solving large instances.

\bibliographystyle{IEEEtran}
\bibliography{main}
\end{document}